\documentclass[10pt,twocolumn,letterpaper]{article}

\usepackage[pagenumbers]{cvpr} 

\usepackage[dvipsnames]{xcolor}

\definecolor{cvprblue}{rgb}{0.21,0.49,0.74}
\usepackage[accsupp]{axessibility} 
\usepackage[pagebackref,breaklinks,colorlinks,citecolor=cvprblue]{hyperref}
\usepackage{multirow}
\usepackage{array}
\usepackage{booktabs}
\newcolumntype{L}[1]{>{\raggedright\arraybackslash}m{#1}}
\newcolumntype{C}[1]{>{\centering\arraybackslash}m{#1}}
\newcolumntype{R}[1]{>{\raggedleft\arraybackslash}m{#1}}
\newcolumntype{+}{>{\global\let\currentrowstyle\relax}}
\newcolumntype{^}{>{\currentrowstyle}}

\newcommand{\RomNum}[1]{\MakeUppercase{\romannumeral #1}}

\def\paperID{12820} 
\def\confName{CVPR}
\def\confYear{2024}

\title{MFP: Making Full Use of Probability Maps for Interactive Image Segmentation}

\author{Chaewon Lee\\
Korea University\\
{\tt\small chaewonlee@mcl.korea.ac.kr}
\and
Seon-Ho Lee\\
Korea University\\
{\tt\small seonholee@mcl.korea.ac.kr}
\and
Chang-Su Kim\thanks{Corresponding author}\\
Korea University\\
{\tt\small changsukim@korea.ac.kr}
}

\begin{document}
\maketitle
\begin{abstract}
In recent interactive segmentation algorithms, previous probability maps are used as network input to help predictions in the current segmentation round. However, despite the utilization of previous masks, useful information contained in the probability maps is not well propagated to the current predictions. In this paper, to overcome this limitation, we propose a novel and effective algorithm for click-based interactive image segmentation, called MFP, which attempts to make full use of probability maps. We first modulate previous probability maps to enhance their representations of user-specified objects. Then, we feed the modulated probability maps as additional input to the segmentation network. We implement the proposed MFP algorithm based on the ResNet-34, HRNet-18, and ViT-B backbones and assess the performance extensively on various datasets. It is demonstrated that MFP meaningfully outperforms the existing algorithms using identical backbones. The source codes are available at \href{https://github.com/cwlee00/MFP}{https://github.com/cwlee00/MFP}.
\end{abstract}
\section{Introduction}
\label{sec:intro}

Interactive image segmentation is a task that aims to segment objects of interest given guidance through user annotations. It enables users to select objects and delineate them easily, so it is useful in many applications such as image editing and medical image analysis. With the rapid development of deep-learning-based algorithms for dense prediction tasks, demands for annotated data have increased significantly. However, obtaining pixel-level annotations is costly due to its laborious and time-consuming traits. With the employment of interactive segmentation techniques, these labeling costs could be reduced. It is hence essential to develop an effective interactive segmentation algorithm.

Various forms of user annotations have been adopted in interactive image segmentation, including bounding boxes \cite{lempitsky2009image,rother2004grabcut}, scribbles \cite{grady2006random, bai2014error}, and clicks \cite{xu2016deep, li2018interactive, jang2019interactive, sofiiuk2020f, chen2021conditional, sofiiuk2022reviving, liu2022pseudoclick, lin2022focuscut, du2023efficient, liu2023simpleclick}. However, click-based interactions have become the mainstream methods due to their simplicity and well-established studies. In click-based methods, a user successively places foreground or background clicks to obtain a segmentation mask. Every time the user places a click, the segmentation mask is updated. Then, based on the segmentation mask, the user provides a new click on the mislabeled areas. This is repeatedly performed until a desired result is obtained. In this work, we attempt to develop a novel and more effective framework for click-based interactive image segmentation.

\begin{figure}[t]
  \centering
  \includegraphics[width=1\linewidth]{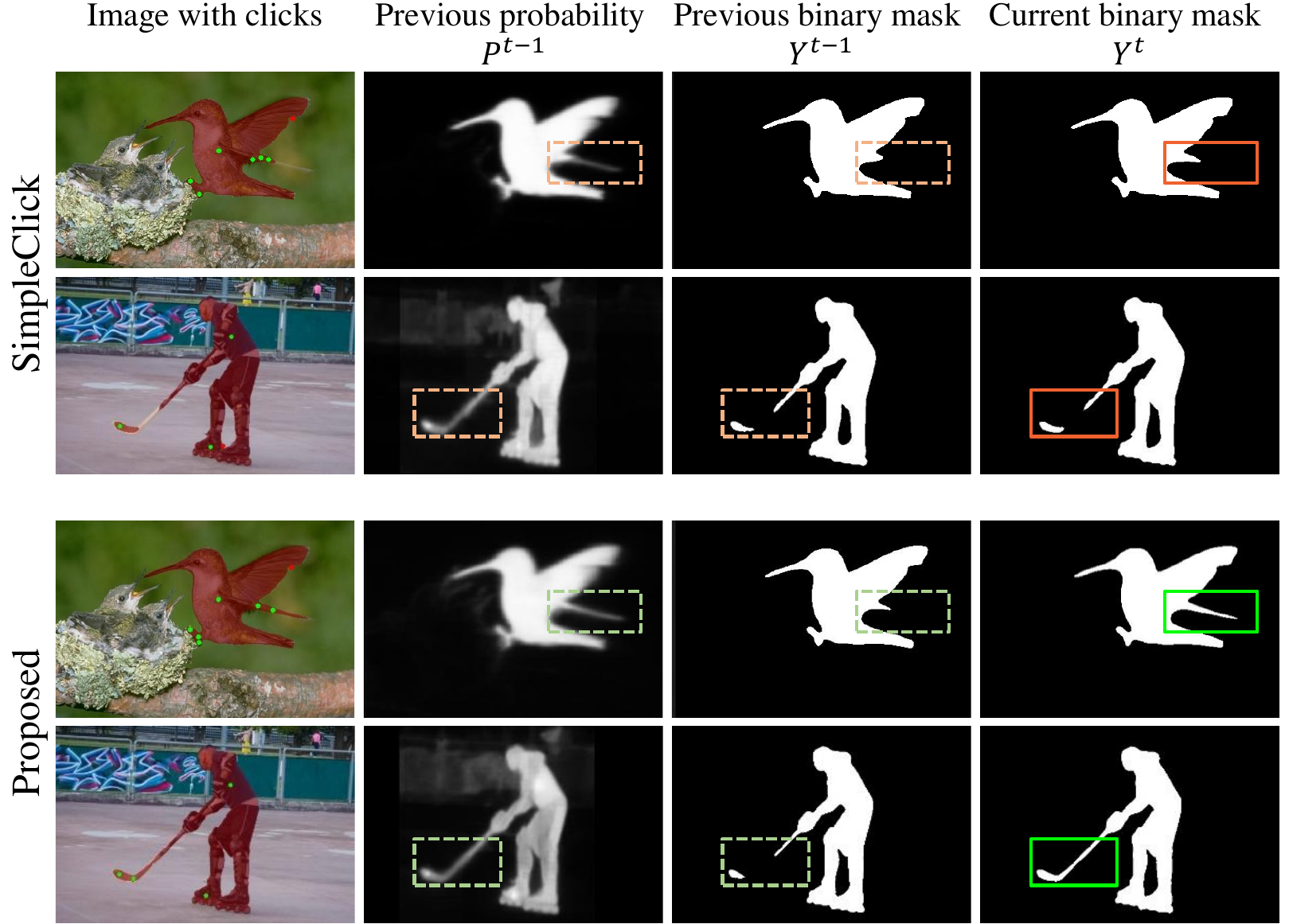}
  \caption{Utilization of previous probability information in the current segmentation round: The conventional click-based interactive segmentation algorithm SimpleClick \cite{liu2023simpleclick} fails to capture the details contained in the previous probability maps. On the other hand, the proposed algorithm exploits the shape details in the previous probability maps to yield a better segmentation result in the current round.}
  \label{fig:opening}
\end{figure}

Recently, deep-learning-based techniques have shown promising results for interactive image segmentation. While early deep methods feed only an input image and click maps to the segmentation networks, Sofiiuk \etal \cite{sofiiuk2022reviving} tried to exploit previous segmentation masks by taking them as additional input to their segmentation network. They demonstrated that making the network model aware of previous masks improves the stability of the model. Since then, taking previous masks (or previous probability maps) as the network input has become a standard pipeline for click-based interactive segmentation. However, even though several algorithms use previous probability maps to generate current predictions, we observe that the information in the previous probability maps is not well propagated to the current predictions. Examples of these observations are presented in Figure~\ref{fig:opening}.

\begin{figure}[t]
  \centering
  \includegraphics[width=1\linewidth]{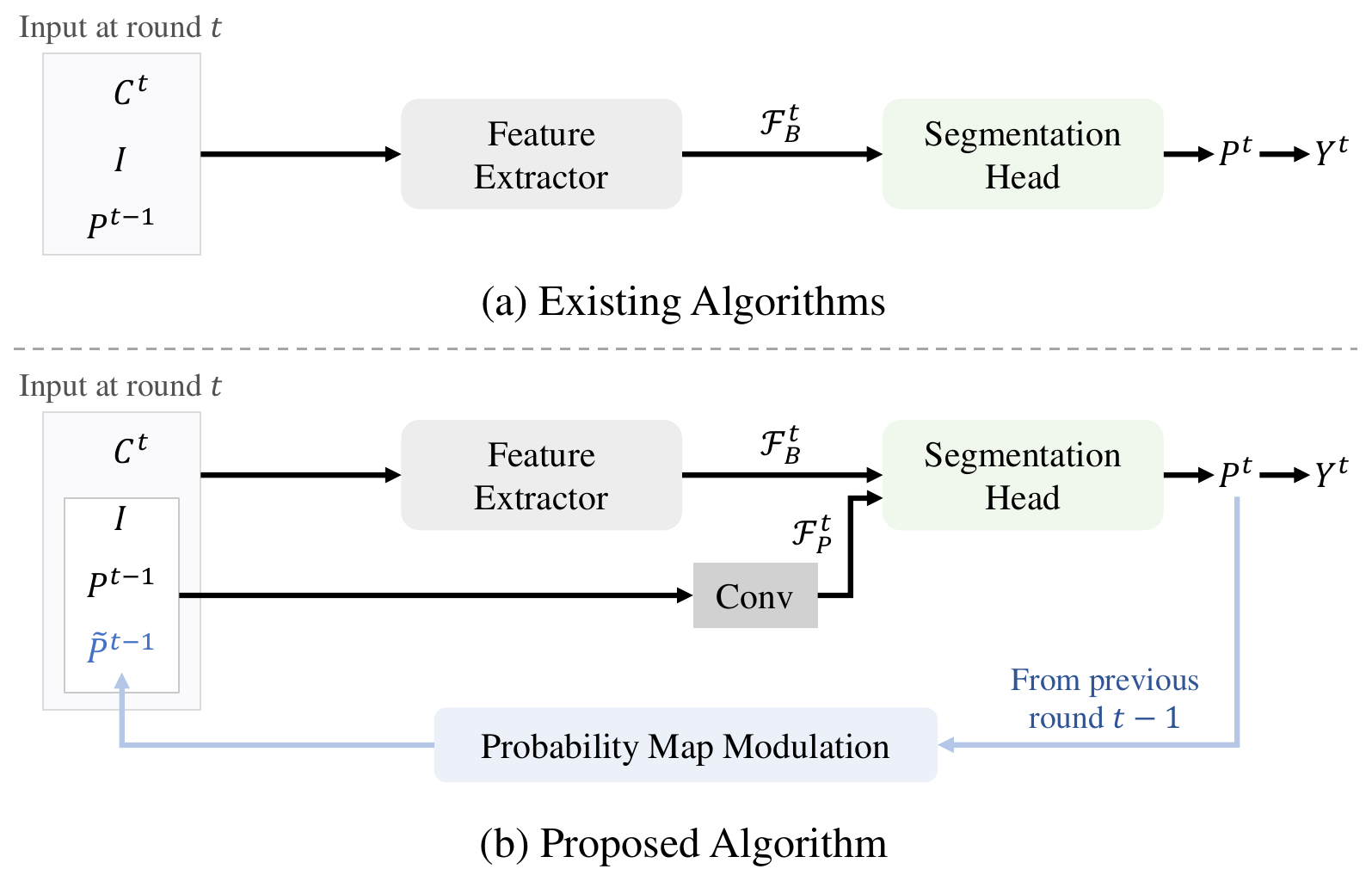}
  \caption{(a) Existing algorithms take an input image $I$, click map $C^{t}$, and previous probability map $P^{t-1}$ as input to the segmentation network. From these input signals, they extract feature ${\cal F}_B^{t}$ and directly feed it into the segmentation head to obtain the current probability map $P^{t}$. Then, $P^{t}$ is thresholded to the final object mask $Y^{t}$. (b) In contrast, the proposed MFP algorithm modulates $P^{t-1}$ into $\tilde{P}^{t-1}$ and takes it as additional input to the network. Furthermore, MFP late-fuses probability-related feature ${\cal F}_P^{t}$ with backbone feature ${\cal F}_B^{t}$ before the segmentation head.}
  \label{fig:network_architecture}
\end{figure}

To overcome these limitations, we propose a novel interactive segmentation framework, called \textbf{M}aking \textbf{F}ull use of \textbf{P}robability maps (MFP), to make better use of previous probabilities. Previous probabilities predicted by a segmentation network provide information such as the shape of a target object, while user clicks give accurate information for discerning foreground regions from background ones. In this paper, we first introduce the notion of probability map modulation to combine these two types of information and yield a better representation of the target object. We propose taking a modulated probability map as additional input to the network. Thus, we extend the existing interactive segmentation framework as shown in Figure~\ref{fig:network_architecture}. Experiments on four benchmark datasets demonstrate that the proposed MFP framework achieves excellent results when implemented on three different backbone networks.

The major contributions of this work can be summarized as follows.
\begin{itemize}
\itemsep0mm
\item We propose the first modulation scheme for previous probability maps that enhances the representation of user-specified objects.

\item We develop the novel MFP framework for click-based interactive segmentation, which propagates click information to unclicked locations effectively by making full use of previous probability maps.

\item We implement the proposed MFP algorithm based on three backbone networks of ResNet-34 \cite{he2016deep}, HRNet-18 \cite{wang2020deep}, and ViT-B \cite{dosovitskiy2021an} and assess the performance on various benchmark datasets. MFP meaningfully outperforms the existing algorithms using identical backbones.
\end{itemize}

\section{Related Work}
\label{sec:related_work}
\subsection{Interactive Image Segmentation}
Extensive research has been carried out to solve the problem of interactive image segmentation. For instance, Rother \etal \cite{rother2004grabcut} proposed an early method, which takes interactive segmentation as a graph-based optimization problem. Such traditional methods rely on handcrafted features, thus they suffer from relatively low performance. With the emergence of deep learning, Xu \etal \cite{xu2016deep} first employed a convolutional neural network to perform interactive segmentation. They encoded user clicks into click maps via the distance transform, and used them with an RGB image as the network input. Since then, their idea of taking click maps as network input has become a de facto standard in deep-learning-based methods \cite{xu2016deep, jang2019interactive, sofiiuk2020f, chen2021conditional}. Although click maps clearly represent the annotated labels in user-clicked locations, the output of the segmentation network is not guaranteed to have correct labels in those locations. To overcome this limitation, Jang and Kim \cite{jang2019interactive} introduced the backpropagating refinement scheme (BRS), which is an inference-time optimization procedure that corrects the mislabeled clicks. Inspired by this work, Sofiiuk \etal \cite{sofiiuk2020f} developed the f-BRS algorithm, which refines features instead of click maps. However, these BRS methods need to run backward gradient passes during the inference, demanding higher computational costs in general.

\begin{figure*}[t]
    \begin{center}
    \includegraphics[width=\linewidth]{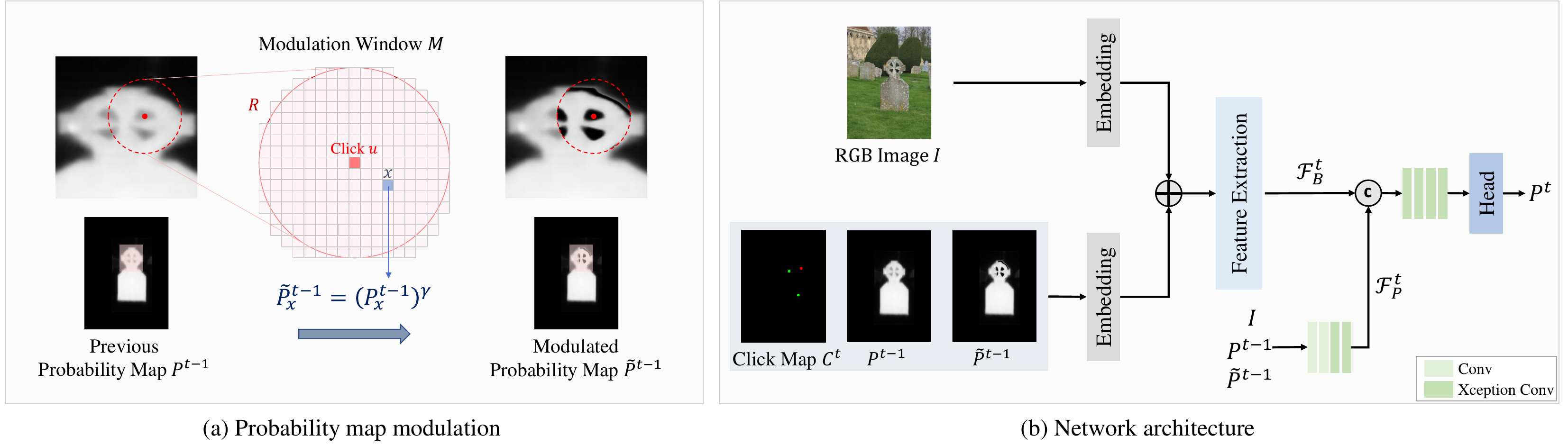}
    \end{center}
    \vspace*{-0.3cm}
    \caption
    {
        Overview of the proposed MFP algorithm. In the click map, foreground and background clicks are depicted in green and red, respectively.
    }
    \label{fig:overview}
\end{figure*}

\subsection{Utilization of Previous Masks}
Forte \etal \cite{forte2020getting} and Sofiiuk \etal \cite{sofiiuk2022reviving} started adding previous segmentation masks as network input in the current segmentation round. In particular, Sofiiuk \etal \cite{sofiiuk2022reviving} demonstrated that even without additional optimization schemes, a simple feed-forward model using previous masks could achieve promising results. After their work, recent interactive segmentation methods in \cite{chen2022focalclick, lin2022focuscut, zhou2023interactive, du2023efficient, liu2023simpleclick} all use the segmentation masks in previous segmentation rounds as network input. To further improve the segmentation performance, Chen \etal \cite{chen2022focalclick} and Lin \etal \cite{lin2022focuscut} proposed to refine segmentation results within a local window. After obtaining a global prediction, their methods also predict local results on cropped image regions around clicks and use the local predictions to refine the global prediction. Zhou \etal \cite{zhou2023interactive} formulated the problem of interactive segmentation as Gaussian process classification, and Du \etal \cite{du2023efficient} proposed to use self-attention and correlation modules for propagating click information to unclicked locations. Liu \etal \cite{liu2023simpleclick} focused on developing a more effective backbone network, and leveraged a plain ViT backbone that could benefit from pretrained weights. Although all these methods use previous masks (or previous probability maps) in the current segmentation round, none of them attempt to explicitly extract beneficial information from the probability maps. Thus, they do not use the previous probability maps to their full potential.

\section{Proposed Algorithm}
\label{sec:proposed_algorithm}
Figure~\ref{fig:overview} is an overview of the proposed MFP algorithm. In each interactive segmentation round, we first modulate the probability map, predicted from the previous clicks, to make it closer to the actual segmentation result the user desires, as illustrated in Figure~\ref{fig:overview}(a). Then, we feed the modulated map as additional input to the segmentation network in Figure~\ref{fig:overview}(b). We train the segmentation network using a recursive training scheme.

\subsection{Probability Map Modulation}

To fully exploit the useful information in the previous prediction, we enhance the shape details of a target object in the probability map $P^{t-1}$ from the previous round $t-1$. For regions that are likely to correspond to the target object, we enhance the probability values to make them closer to the foreground label of 1. On the contrary, for background regions, we lower the probabilities to make them closer to the background label of 0. To this end, we use the method of gamma correction \cite{gonzalez2008digital}.

\vspace*{0.15cm}
\noindent\textbf{Gamma correction:}
In round $t$, we use the probability map $P^{t-1}$ that the segmentation head yielded in the last round $t-1$. We first modulate $P^{t-1}$ into $\tilde{P}^{t-1}$ via gamma correction.
We modify only the region that seems to need further refinement. More specifically, we apply gamma correction only to the pixels within a modulation window $M$, as shown in Figure~\ref{fig:overview}(a). Typically, pixels nearer to the current click $u$ are more likely to require further refinement. Hence, we define the modulation window as
\begin{equation}
M = \{x: \|x-u\|\leq R \}
\label{eq:modulation}
\end{equation}
where $R$ is the radius of the window, representing the attention scope of the current click $u$.

\begin{figure}[t]
  \centering
  \includegraphics[width=1\linewidth]{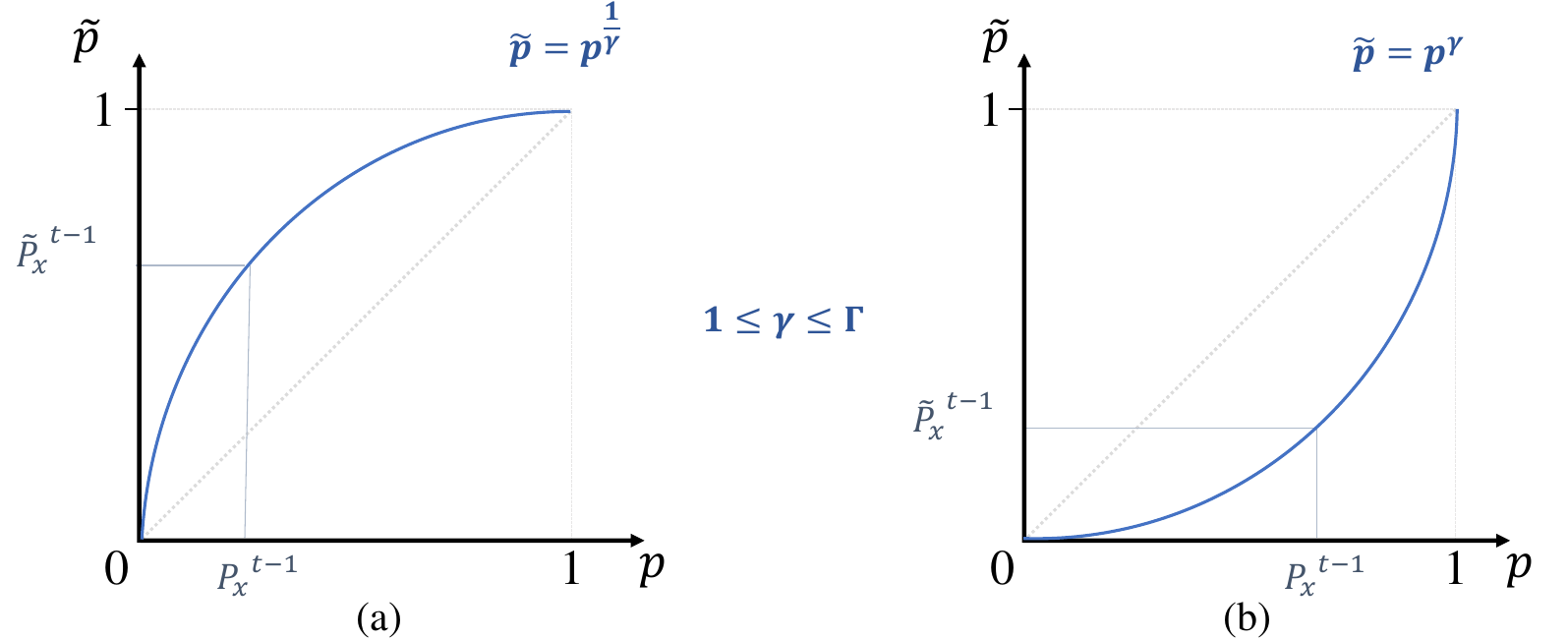}
  \caption{Illustration of the gamma correction for the probability modulation: (a) probabilities near a foreground click are increased with power $\frac{1}{\gamma}$ smaller than 1, and (b) those near a background click are decreased with power $\gamma$ bigger than 1. }
  \label{fig:gamma_correction}
\end{figure}

Suppose that pixel $x$ belongs to the modulation window, \ie $x \in M$. Also, let $P_x$ denote the probability of $x$ in a probability map $P$. Then, $P_x^{t-1}$ is modulated to $\tilde{P}_x^{t-1}$ by gamma correction. Let $l(u)$ be the label that the user provides at the current click $u$. Note that user clicks are provided on mislabeled areas. Therefore, if a foreground click is given with $l(u) = 1$, it indicates that a background label was assigned to $u$ in the last round and $P^{t-1}_u < 0.5$.
Since the user intends to assign a foreground label to $u$, we increase the probabilities of pixels in $M$. Therefore, when $l(u) = 1$, we perform the increasing modulation
\begin{equation}\label{eq:gamma_correction_foreground}
    \tilde{P}_x^{t-1} = (P_x^{t-1})^{\frac{1}{\gamma}}
\end{equation}
for each $x\in M$, where $1 \leq \gamma \leq \Gamma$. Figure~\ref{fig:gamma_correction}(a) shows a gamma correction curve for the increasing modulation. The exact opposite processing is done for a background click. When $l(u) = 0$, we perform the decreasing modulation
\begin{equation}\label{eq:gamma_correction_background}
    \tilde{P}_x^{t-1} = (P_x^{t-1})^\gamma
\end{equation}
for each $x\in M$, as shown in Figure~\ref{fig:gamma_correction}(b).

\vspace*{0.15cm}
\noindent\textbf{Assignment of $\gamma$:}
We assign different values of gamma for modulating each pixel $x$ in $M$. The assignment is done according to how far $x$ is from the given click $u$. For the click itself, the desired value $\tilde{P}_u^{t-1}$ is clear; it is desired that $\tilde{P}^{t-1}_u \approx 1$ if $l(u) = 1$, and $\tilde{P}^{t-1}_u \approx 0$ if $l(u) = 0$. It is hence natural to assign the biggest gamma $\Gamma$ to the current click $u$. We set $\Gamma$ so that $\tilde{P}^{t-1}_u = 0.99$ for a foreground click and $\tilde{P}^{t-1}_u = 0.01$ for a background click. On the other hand, pixels far away from $u$ are less likely to belong to the same object as $u$, so we assign smaller gamma values for those pixels. For measuring how far a pixel is from the click, we propose two distance metrics: Euclidean distance and probability distance. We also develop different gamma assignment schemes for the two distance metrics. Table~\ref{table:modulation_schemes} summarizes these two schemes.

\begin{table}[t]\centering
    \renewcommand{\arraystretch}{1.2}
    \caption
    {
        Two gamma assignment schemes.
    }
    \vspace*{-0.15cm}
    \resizebox{1.0\linewidth}{!}{
    \begin{tabular}[t]{+L{1.5cm}^L{2.9cm}^L{5cm}}
    \toprule
    Method & Distance metric & Gamma assignment scheme \\
    \midrule
        Euclidean distance    & $d =\|x - u\|$ & $\gamma = \Gamma \cdot (1 - \frac{d}{R}) + \frac{d}{R}$ \\
    \midrule
        Probability distance & $d =(P^{t-1}_x - P^{t-1}_u)^2$ & $\gamma = (\Gamma-1) \cdot \max \left\{\frac{(\bar{d}-d)^3}{\bar{d}^3} , 0 \right\} + 1$ \\
    \bottomrule
    \end{tabular}}
    \vspace{-0.2cm}
    \label{table:modulation_schemes}
\end{table}

First, we measure how far pixel $x$ is from click $u$ by the Euclidean distance $d=\|x-u\|$. In this case, $\gamma$ is determined by
\begin{equation}\label{eq:gamma_physical}
    \gamma = \Gamma\cdot(1 - \frac{d}{R}) + \frac{d}{R}.
\end{equation}
Thus, $\gamma$ linearly decreases from $\Gamma$ to 0, as pixel $x$ moves away from click $u$ to the boundary of the modulation window $M$ in \eqref{eq:modulation}.

Second, assuming that similar pixels in the same region would be assigned similar probabilities by the segmentation network, we define the probability distance $d =(P^{t-1}_x - P^{t-1}_u)^2$ between $x$ and $u$. In other words, pixel $x$ is considered farther from click $u$ when its probability $P^{t-1}_x$ differs more from $P^{t-1}_u$. In this case, the maximum distance cannot be bounded by the radius $R$ of the modulation window $M$. Instead, we measure the probability distances of all pixels in $M$ from click $u$. Let $\bar{d}$ denote the median of these distances. Then, we determine $\gamma$ by
\begin{equation}\label{eq:gamma_intensity}
   \gamma = (\Gamma-1) \cdot \frac{(\bar{d}-d)^3}{{\bar{d}}^3} + 1
\end{equation}
when $d \leq \bar{d}$. Therefore, $\gamma = \Gamma$ at click $u$, and $\gamma = 1$ when $d=\bar{d}$. When $d> \bar{d}$, we assume that $x$ and $u$ are unlikely to belong to the same region and set $\gamma = 1$.

\begin{figure}[t]
  \centering
  \includegraphics[width=1\linewidth]{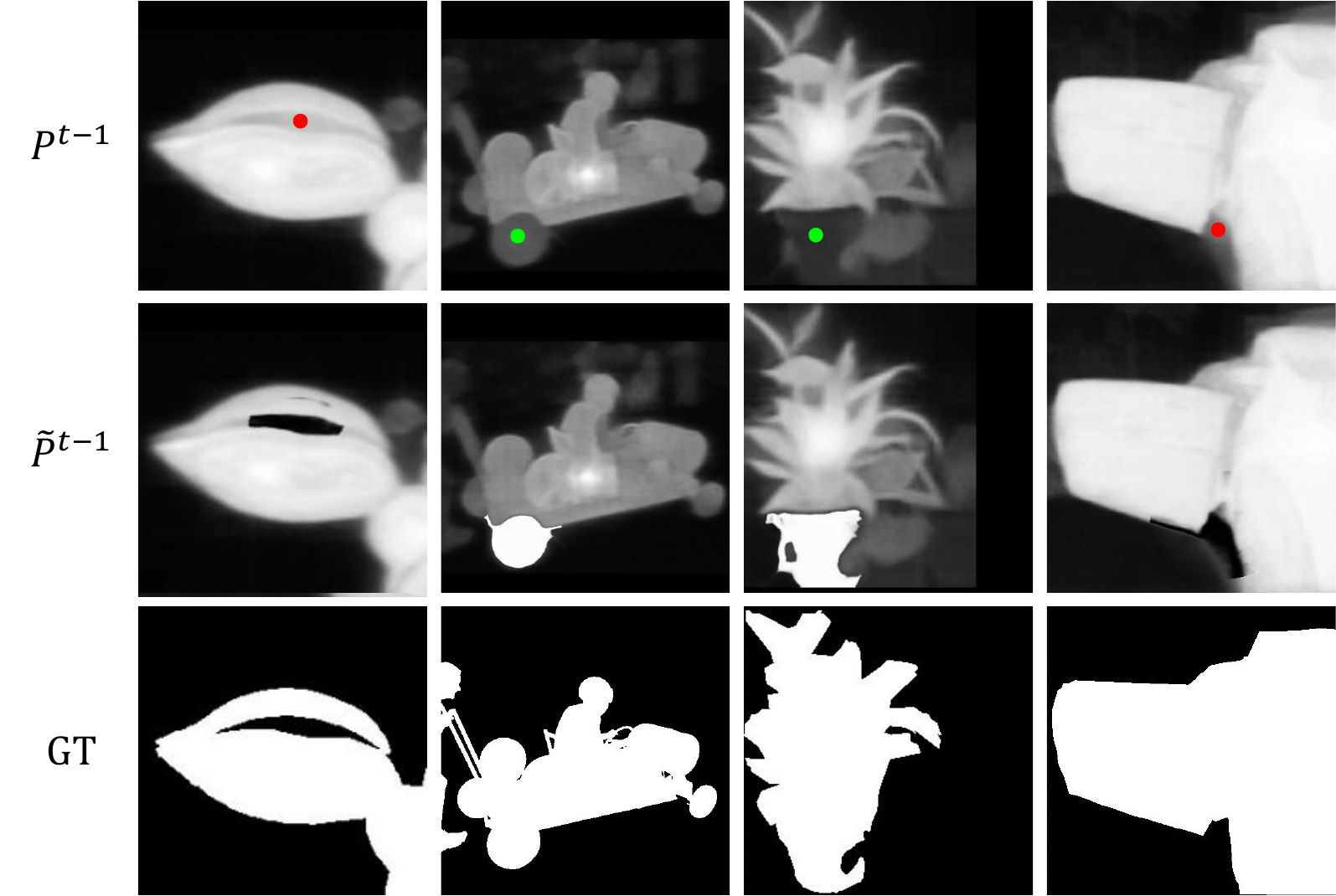}
  \caption{Examples of the probability map modulation via \eqref{eq:gamma_intensity}. From top to bottom: previous probability maps ${P}^{t-1}$ before modulation with current clicks, modulated probability maps $\tilde{P}^{t-1}$, and ground-truth masks of target objects.}
  \label{fig:prevMod_examples}
\end{figure}

We adopt the probability-distance-based scheme in \eqref{eq:gamma_intensity} for early clicks up to the $N$th click and the Euclidean-distance-based scheme in \eqref{eq:gamma_physical} for later clicks. This is because, in early rounds, rough and large-scale object shapes tend to remain in $P^{t-1}$. On the other hand, in late rounds, the distinction between the target object and the background becomes clearer, and fine-scale information is required for better segmentation. However, $P^{t-1}$ does not contain such fine-scale information in general. Thus, the Euclidean distance is used instead in late rounds. Figure~\ref{fig:prevMod_examples} shows examples of the probability map modulation via the gamma assignment scheme in \eqref{eq:gamma_intensity}.

\vspace*{0.15cm}
\noindent\textbf{Assignment of $R$:}
We set the radius $R$ of $M$ to be $R_{\max}$. However, when there are previous clicks of the opposite type, we set $R$ to be half the minimum distance of the current click $u$ to an opposite click. Thus, we set
\begin{equation}
   R = \min \left\{\frac{1}{2} \cdot \min_{c\in O} \|u - c\|, \, R_{\max} \right\}
\end{equation}
where O is the set of previous clicks of the opposite type.

\begin{table*}[t]\centering
    \vspace*{-0.2cm}
    \renewcommand{\arraystretch}{1.1}
    \caption
    {
        The NoC scores of the proposed MFP algorithm and existing algorithms, which are trained on the SBD dataset \cite{hariharan2011semantic}.
    }
    \vspace*{-0.15cm}
    \resizebox{1.0\linewidth}{!}{
    \begin{tabular}[t]{+L{2.8cm}^L{1.7cm}^C{1.0cm}^C{1.0cm}^C{1.0cm}^C{1.0cm}^C{1.0cm}^C{1.0cm}^C{1.0cm}^C{1.0cm}^C{1.0cm}^C{1.0cm}^C{1.0cm}^C{1.0cm}}
    \toprule
     \multirow{2}[2]{*}{} & \multirow{2}[2]{*}{} & \multicolumn{3}{c}{GrabCut} &  \multicolumn{3}{c}{Berkeley} & \multicolumn{3}{c}{DAVIS} & \multicolumn{3}{c}{SBD}\\
     \cmidrule(lr){3-5} \cmidrule(lr){6-8} \cmidrule(lr){9-11} \cmidrule(lr){12-14}
    Algorithm & Backbone & NoC85 & NoC90& NoC95& NoC85 & NoC90& NoC95& NoC85 & NoC90& NoC95& NoC85 & NoC90& NoC95\\
    \midrule
         BRS \cite{jang2019interactive} &  DenseNet & 2.60 & 3.60 & - & - & 5.08 & - & 5.58 & 8.24 & - & 6.59 & 9.78 & - \\
         f-BRS \cite{sofiiuk2020f} &  ResNet-101 & 2.30 & 2.72 & - & - & 4.57 & - & 5.04 & 7.41 & - & 4.81 & 7.73 & - \\
         RITM \cite{sofiiuk2022reviving} &  HRNet-18 & 1.76 & 2.04 & 3.66 & 1.87 & 3.22 & 8.35 & 4.94 & 6.71 & 13.87 & 3.39 & 5.43 & 11.65\\
         CDNet \cite{chen2021conditional} &  ResNet-34 & 1.86 & 2.18 & 3.68 & 1.95 & 3.27 & 8.29 & 5.00 & 6.89 & 14.24 & 5.18 & 7.89 & 14.27 \\
         PsuedoClick \cite{liu2022pseudoclick} &  HRNet-18 & 1.68 & 2.04 & - & 1.85 & 3.23 & - & 4.81 & 6.57 & - & 3.38 & 5.40 & - \\
         FocalClick \cite{chen2022focalclick} &  SegF-B0 & 1.66 & 1.90 & - & - & 3.14 & - & 5.02 & 7.06 & - & 4.34 & 6.51 & -\\
         FocusCut \cite{lin2022focuscut} &  ResNet-101 & 1.46 & 1.64 & - & 1.81 & 3.01 & - & 4.85 & 6.22 & - & 3.40 & \underline{5.31} & -\\
         EMC-Click \cite{du2023efficient} & HRNet-18 & 1.74 & 1.84 & - & - & 3.03 & - & 5.05 & 6.71 & - & 3.38 & 5.51 & -\\
         GPCIS \cite{zhou2023interactive} & ResNet-50 & 1.64 & 1.82 & 2.62 & 1.60 & 2.60 & 6.77 & 4.37 & 5.89 & 12.42 & 3.80 & 5.71 & \bf{11.06}  \\
         iCMFormer \cite{li2023interactive} & ViT-B & \bf{1.36} & \bf{1.42} & - & \underline{1.42} & 2.52 & - & \underline{4.05} & 5.58 & - & 3.33 & \underline{5.31} & - \\
         SimpleClick \cite{liu2023simpleclick} & ViT-B & 1.40 & 1.54 & \underline{2.16} & 1.44 & \underline{2.46} & \underline{6.70} & 4.10 & \underline{5.48} & \underline{12.24} & \underline{3.28} & \bf{5.24} & 11.24 \\
    \midrule
         MFP {\scriptsize (Proposed)} & ViT-B & \underline{1.38} & \underline{1.48} & \bf{1.92}  & \bf{1.39} &\bf{2.17} & \bf{6.18} & \bf{3.92}& \bf{5.32} & \bf{11.27} & \bf{3.21}& \bf{5.24} & \underline{11.20}  \\
    \bottomrule
    \end{tabular}}
    \label{table:comparative_assessment_sbd}
    \vspace*{-0.2cm}
\end{table*}

\subsection{Network Architecture}

After obtaining the modulated probability map $\tilde{P}^{t-1}$, we feed it together with the input image $I$, click map $C^t$, and original probability map $P^{t-1}$ into the segmentation network, as depicted in Figure~\ref{fig:overview}(b). For the interactive segmentation task, we adopt a common semantic segmentation network as the backbone for feature extraction. However, the semantic segmentation network takes only an RGB image as input. To handle the additional input $C^{t}$, $P^{t-1}$, and $\tilde{P}^{t-1}$, we concatenate them, and embed the concatenation and the input image $I$, respectively, into tensors of the same size, as done in \cite{sofiiuk2022reviving,liu2023simpleclick}. Then, we element-wise add these tensors and convey the sum to the feature extraction network. As the feature extraction backbone, we test ResNet-34 \cite{he2016deep}, HRNet-18 \cite{wang2020deep}, and ViT-B \cite{dosovitskiy2021an} of different complexities, which are widely used for interactive image segmentation. The feature extractor yields a feature map ${\cal F}_B^{t}$.

Existing interactive segmentation algorithms directly input the backbone feature ${\cal F}_B^{t}$ into a segmentation head to obtain a segmentation result. In contrast, we propose fusing the probability maps $P^{t-1}$ and $\tilde{P}^{t-1}$ with ${\cal F}_B^{t}$. This late fusion strengthens the influence of the probability information on the final segmentation result. Specifically, we first concatenate $I$, ${P}^{t-1}$, $\tilde{P}^{t-1}$ and adjust their spatial resolutions and channel sizes using two convolution blocks. We then use two Xception conv blocks \cite{chollet2017xception} to extract the probability-related feature ${\cal F}_P^{t}$. Then, we concatenate ${\cal F}_P^{t}$ with ${\cal F}_B^{t}$ and fuse them through four Xception conv blocks. The segmentation head processes this fused feature to generate the probability map $P^t$ in the current round $t$. Thresholding this map yields the final segmentation mask $Y^{t}$.

\subsection{Recursive Training}
As the proposed MFP algorithm uses ${P}^{t-1}$ to predict ${P}^{t}$, its training requires an ordered series of clicks and the corresponding probability maps. Sofiiuk \etal \cite{sofiiuk2022reviving}, one of the first methods to use previous probability maps in interactive segmentation, adopt an iterative sampling strategy. They first sample user clicks randomly, as done in \cite{xu2016deep}. After the random clicks, they add $0 \sim 3$ clicks iteratively based on the errors in the network prediction results. Although they partly train the model by mimicking user interactions, they still resort to random sampling initially. Thus, for initial clicks, there are no previous probability maps, and the probability modulation cannot be performed. Therefore, in this work, we develop a fully recursive training strategy to exploit the information in the ordered series of interactive clicks. More specifically, we start the training of an image by sampling the first click near the center of a target object. Then, by comparing the result of the segmentation network with the ground truth, we select the next click near the center of the biggest error region. This recursive selection and training is performed up to 24 clicks for an image.

\begin{figure*}[t]
    \begin{center}
    \includegraphics[width=\linewidth]{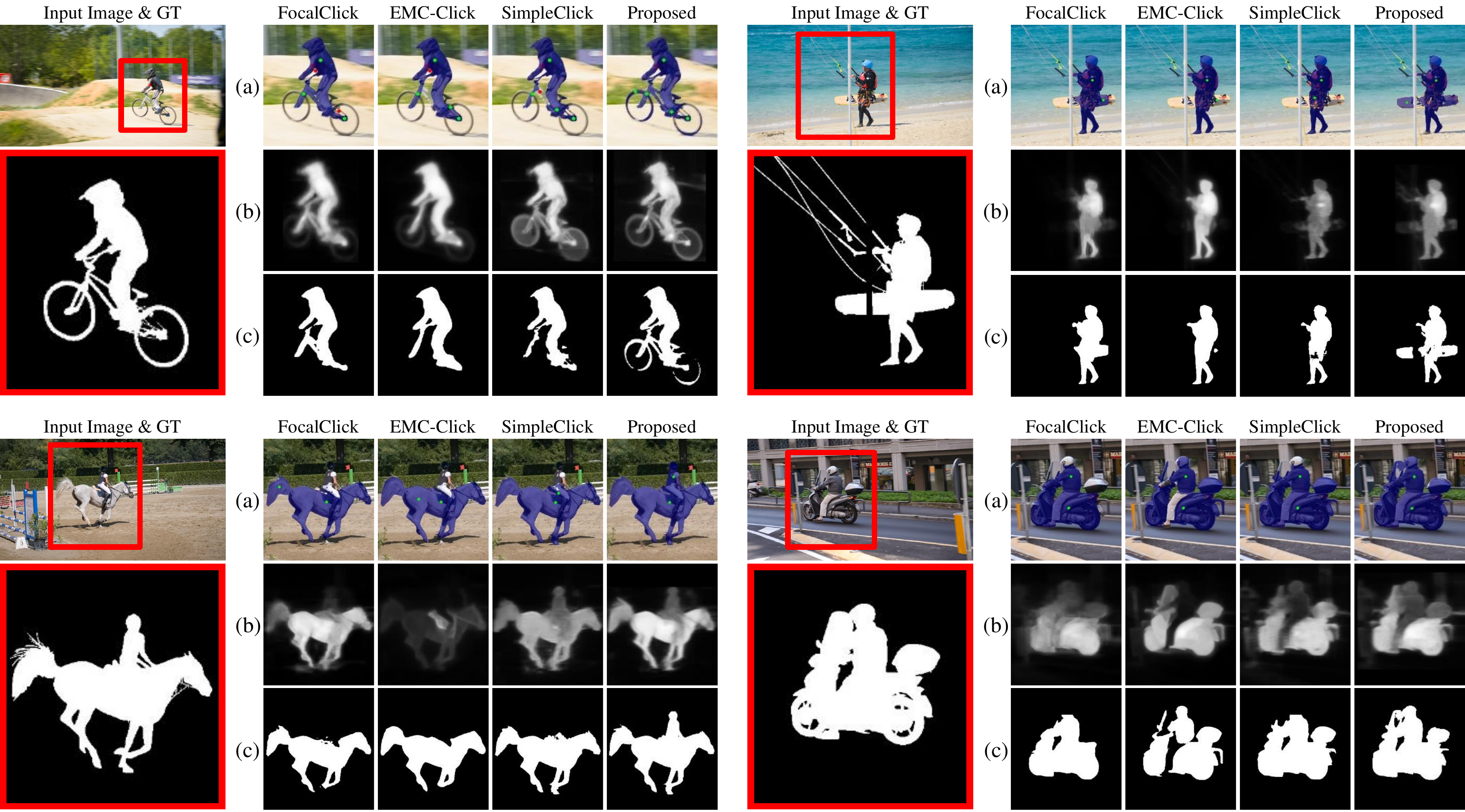}
    \end{center}
    \vspace*{-0.5cm}
    \caption
    {
        Qualitative comparison of algorithms trained on the COCO+LVIS dataset. We compare the proposed MFP with conventional algorithms that employ previous masks as input: FocalClick \cite{chen2022focalclick}, EMC-Click \cite{du2023efficient}, and SimpleClick \cite{liu2023simpleclick}. Rows (a) show input images with clicks and prediction masks overlayed. Rows (b) and (c) show previous probability maps and current prediction masks, respectively. Note that the algorithms are compared fairly using the same automatic clicking strategy. Since the algorithms produce different segmentation results, click locations may differ accordingly.
    }
    \label{fig:qualitative_comparison}
    \vspace*{-0.25cm}
\end{figure*}

\section{Experiments}
\label{sec:experiments}

\subsection{Experimental Settings}
\noindent\textbf{Datasets:}
We use four widely used datasets GrabCut \cite{rother2004grabcut}, Berkeley \cite{mcguinness2010comparative}, DAVIS \cite{perazzi2016benchmark}, and SBD \cite{hariharan2011semantic} to assess the proposed MFP algorithm. SBD and a combination of datasets COCO \cite{lin2014microsoft} and LVIS\cite{gupta2019lvis} are used for training.

\vspace*{0.1cm}
\begin{itemize}
\itemsep1mm
\item GrabCut: It consists of 50 images. For each image, a single object mask is provided.
\item Berkeley: It contains 96 images with 100 object masks.
\item DAVIS: Although constructed for video object segmentation, it can be also used for interactive image segmentation by sampling frames from the videos. We use the same 345 frames, sampled from 50 videos, as in \cite{jang2019interactive}.
\item SBD: It is divided into training and validation sets. The training set has 8,498 images with 20,172 instance masks, and the validation set has 2,857 images with 6,671 masks.
\item COCO + LVIS: COCO consists of 99K images with 1.2M instance masks, and LVIS has 100K images with 1.2M masks. As in \cite{sofiiuk2022reviving}, we combine these two datasets and use 104K images with 1.6M instance masks for training.
\end{itemize}

\vspace*{0.1cm}
\noindent\textbf{Evaluation Metrics:}
To evaluate the proposed algorithm, we use two performance measures. First, we report the NoC score, which is the average number of clicks required to achieve a certain intersection-over-union (IoU) ratio. Previous studies usually set the target IoU ratio as 90\% and report NoC@85 and NoC@90 for assessment. However, recent methods yield high-quality segmentation results. Therefore, we set the target IoU score as 95\% and report NoC@95 additionally. Second, we plot the mean intersection-over-union (mIoU) score according to the number of clicks and report the area under the curve (AUC).

\vspace*{0.1cm}
\noindent\textbf{Implementation Details:}
The proposed MFP can be implemented upon various backbone networks. In this work, we implement three versions based on the ViT-B \cite{dosovitskiy2021an}, HRNet-18 + OCR \cite{wang2020deep, yuan2020object}, and ResNet-34 + DeepLabv3+ \cite{he2016deep,chen2018encoder} backbones. Also, we use the SBD \cite{hariharan2011semantic} and COCO + LVIS \cite{lin2014microsoft, gupta2019lvis} datasets for training. We apply random resizing, cropping, flipping, rotation, and brightness control for data augmentation. We minimize the normalized focal loss \cite{sofiiuk2022reviving} using the Adam optimizer with $\beta_1=0.9$ and $\beta_2=0.999$. We fix the hyperparameters in the probability map modulation to $N=7$ and $R_{\max}=100$ in all experiments. For evaluation, results can be greatly affected by how a user places clicks. Thus, for reliable assessment, we adopt the automatic clicking strategy used in \cite{xu2016deep, jang2019interactive, chen2022focalclick}.

\begin{table*}[t]\centering
    \renewcommand{\arraystretch}{1.1}
    \caption
    {
        The NoC scores of the proposed MFP algorithm and existing algorithms, which are trained on the COCO + LVIS datasets \cite{lin2014microsoft, gupta2019lvis}.
    }
    \vspace*{-0.15cm}
    \resizebox{1.0\linewidth}{!}{
    \begin{tabular}[t]{+L{2.8cm}^L{1.7cm}^C{1.0cm}^C{1.0cm}^C{1.0cm}^C{1.0cm}^C{1.0cm}^C{1.0cm}^C{1.0cm}^C{1.0cm}^C{1.0cm}^C{1.0cm}^C{1.0cm}^C{1.0cm}}
    \toprule
     \multirow{2}[2]{*}{} & \multirow{2}[2]{*}{} & \multicolumn{3}{c}{GrabCut} &  \multicolumn{3}{c}{Berkeley} & \multicolumn{3}{c}{DAVIS} & \multicolumn{3}{c}{SBD}\\
     \cmidrule(lr){3-5} \cmidrule(lr){6-8} \cmidrule(lr){9-11} \cmidrule(lr){12-14}
    Algorithm& Backbone& NoC85 & NoC90& NoC95& NoC85 & NoC90& NoC95& NoC85 & NoC90& NoC95& NoC85 & NoC90& NoC95\\
    \midrule
         RITM \cite{sofiiuk2022reviving} &  HRNet-32 & 1.46 & 1.56 & 2.48 & 1.43 & 2.10 & 5.41 & 4.11 & 5.34 & 11.51 & 3.59 & 5.71 & 12.00 \\
         CDNet \cite{chen2021conditional} &  ResNet-34 & 1.40 & 1.52 & 1.84 & 1.47 & 2.06 & 5.42 & 4.27 & 5.56 & 11.90 & 4.30 & 7.04 & 14.17 \\
         PseudoClick \cite{liu2022pseudoclick} &  HRNet-32 & 1.36 & 1.50 & - & 1.40 & 2.08 & - & 3.79 & 5.11 & - & 3.46 & 5.54 & -\\
         FocalClick \cite{chen2022focalclick} &  SegF-B3 & 1.44 & 1.50 & 1.82 & 1.55 & 1.92 & \bf{4.63} & 3.61 & \underline{4.90} & 10.58 & 3.43 & 5.62 & \bf{11.55} \\
         EMC-Click \cite{du2023efficient} & HRNet-32 & \bf{1.30} & \bf{1.42} & 1.84 & 1.48 & 2.35 & 6.95 & 4.29 & 5.33 & 11.82 & 3.55 & 5.65 & 12.26\\
         DynaMITe \cite{rana2023dynamite} & Swin-L & 1.62 & 1.72 & - & 1.39 & \underline{1.90} & - & 3.80 & 5.09 & - & 3.32 & 5.64 & - \\
         iCMFormer \cite{li2023interactive} & ViT-B & 1.42 & 1.52 & - & 1.40 & \bf{1.86} & - & \underline{3.40} & 5.06 & - & \underline{3.29} & \bf{5.30} & - \\
         SimpleClick \cite{liu2023simpleclick} & ViT-B & 1.38 & \underline{1.48} & \underline{1.80} & \underline{1.36} & 1.97 & 5.05 & 3.66 & 5.06 & \underline{10.04} & 3.43 & 5.62 & 11.92 \\
    \midrule
         MFP {\scriptsize (Proposed)} & ViT-B & \underline{1.34} & \bf{1.42} & \bf{1.70}  & \bf{1.35} &\underline{1.90} & \underline{4.68} & \bf{3.37}& \bf{4.81} & \bf{9.23} & \bf{3.26}& \underline{5.34} & \underline{11.65}  \\
    \bottomrule
    \end{tabular}}
    \label{table:comparative_assessment_coco}
\end{table*}

\begin{table*}[t]\centering
    \renewcommand{\arraystretch}{1.1}
    \caption
    {
        Comparison of the proposed MFP algorithm with conventional algorithms using identical backbones. The algorithms are grouped according to the backbone networks employed. All algorithms are trained on the SBD dataset \cite{hariharan2011semantic}.
    }
    \vspace*{-0.15cm}
    \resizebox{1.0\linewidth}{!}{
    \begin{tabular}[t]{+L{2.8cm}^L{1.7cm}^C{1.0cm}^C{1.0cm}^C{1.0cm}^C{1.0cm}^C{1.0cm}^C{1.0cm}^C{1.0cm}^C{1.0cm}^C{1.0cm}^C{1.0cm}^C{1.0cm}^C{1.0cm}}
    \toprule
     \multirow{2}[2]{*}{} & \multirow{2}[2]{*}{} & \multicolumn{3}{c}{GrabCut} &  \multicolumn{3}{c}{Berkeley} & \multicolumn{3}{c}{DAVIS} & \multicolumn{3}{c}{SBD}\\
     \cmidrule(lr){3-5} \cmidrule(lr){6-8} \cmidrule(lr){9-11} \cmidrule(lr){12-14}

    Algorithm& Backbone& NoC85 & NoC90& NoC95& NoC85 & NoC90& NoC95& NoC85 & NoC90& NoC95& NoC85 & NoC90& NoC95\\
    \midrule
         CDNet \cite{chen2021conditional} &  ResNet-34 & 1.86 & 2.18 & 3.68 & 1.95 & \bf{3.27} & 8.29 & 5.00 & \bf{6.89} & \bf{14.24} & 5.18 & 7.89 & 14.27 \\
         MFP {\scriptsize (Proposed)} & ResNet-34 & \bf{1.70} & \bf{1.92} & \bf{3.00}  & \bf{1.71} & 3.37 & \bf{7.94} & \bf{4.97} & 7.78 & 14.93 & \bf{3.92} & \bf{6.21}  & \bf{12.47} \\
    \midrule
         RITM \cite{sofiiuk2022reviving} &  HRNet-18 & 1.76 & 2.04 & 3.66 & 1.87 & 3.22 & 8.35 & 4.94 & 6.71 & 13.87 & \underline{3.39} & \underline{5.43} & 11.65\\
         PsuedoClick \cite{liu2022pseudoclick} &  HRNet-18 & \underline{1.68} & 2.04 & - & \underline{1.85} & 3.23 & - & \underline{4.81} & \underline{6.57} & - & \bf{3.38} & \bf{5.40} & - \\
         EMC-Click \cite{du2023efficient} & HRNet-18 & 1.74 & \underline{1.84} & - & - & \bf{3.03} & - & 5.05 & 6.71 & - & \bf{3.38} & 5.51 & -\\
         MFP {\scriptsize (Proposed)} & HRNet-18 & \bf{1.52} & \bf{1.60} & \bf{2.90}  & \bf{1.68} & \underline{3.04} & \bf{7.94} & \bf{4.77}& \bf{6.36} & \bf{13.45} & 3.43 & 5.45 & \bf{11.58}  \\

    \midrule
         iCMFormer \cite{li2023interactive} & ViT-B & \bf{1.36} & \bf{1.42} & - & \underline{1.42} & 2.52 & - & \underline{4.05} & 5.58 & - & 3.33 & 5.31 & - \\

         SimpleClick \cite{liu2023simpleclick} & ViT-B & 1.40 & 1.54 & 2.16 & 1.44 & \underline{2.46} & 6.70 & 4.10 & \underline{5.48} & 12.24 & \underline{3.28} & \bf{5.24} & 11.24 \\

         MFP {\scriptsize (Proposed)} & ViT-B & \underline{1.38} & \underline{1.48} & \bf{1.92}  & \bf{1.39} &\bf{2.17} & \bf{6.18} & \bf{3.92}& \bf{5.32} & \bf{11.27} & \bf{3.21}& \bf{5.24} & \bf{11.20}  \\
    \bottomrule
    \end{tabular}}
    \label{table:backbone-comparisons}
    \vspace*{-0.2cm}
\end{table*}

\begin{figure*}[t]
    \vspace*{-0.2cm}
    \begin{center}
    \includegraphics[width=\linewidth]{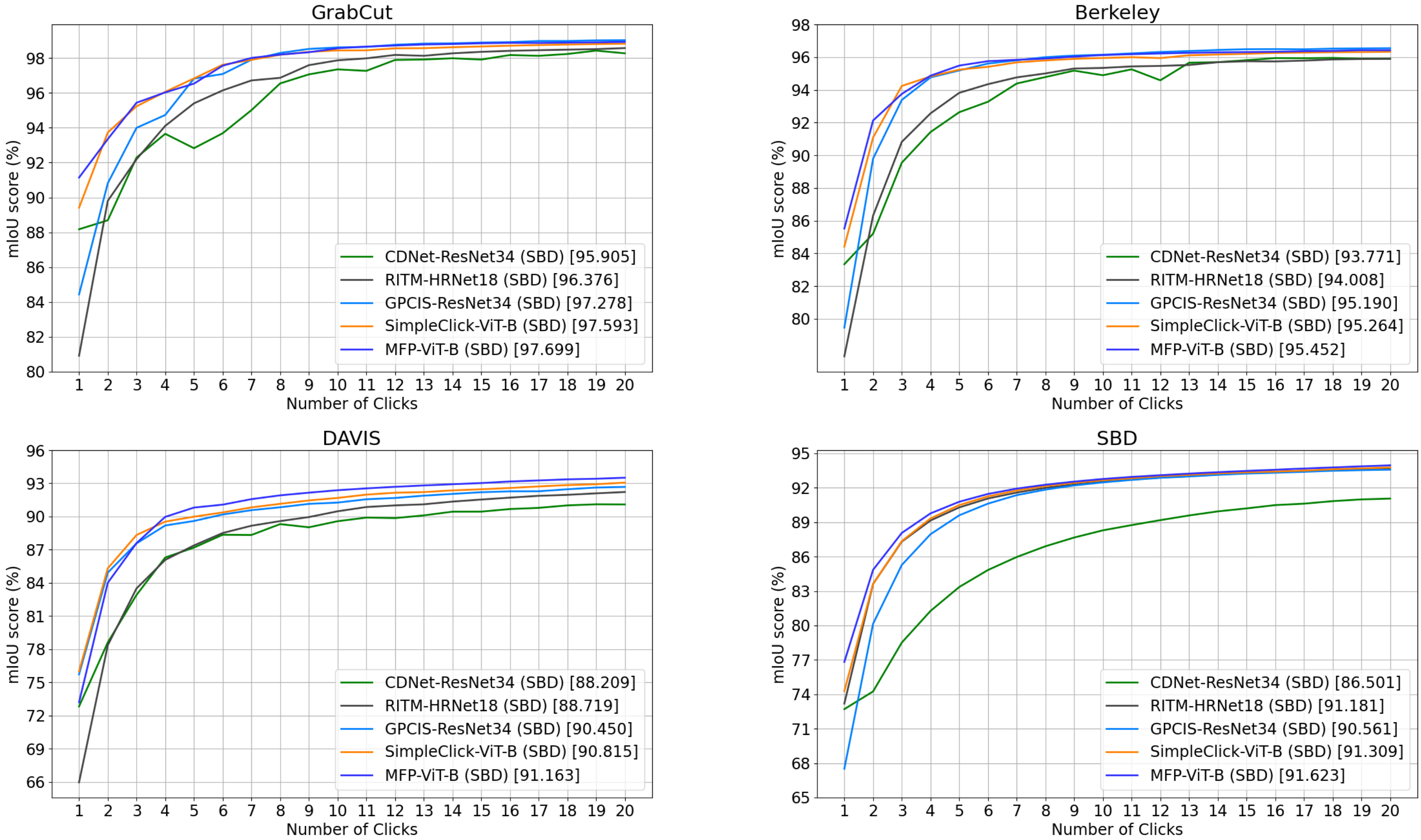}
    \end{center}
    \vspace*{-0.6cm}
    \caption
    {
        Comparison of the mean IoU scores according to the number of clicks on the GrabCut\cite{rother2004grabcut}, Berkeley\cite{mcguinness2010comparative}, DAVIS\cite{perazzi2016benchmark}, and SBD\cite{hariharan2011semantic} datasets. The models are trained on the SBD datasets. The legend of the graph contains the AuC score for each algorithm.
    }
    \label{fig:IoUs}
\end{figure*}

\subsection{Comparative Assessment}

\noindent\textbf{Comparison with state-of-the-art methods:}
We first compare the proposed MFP (ViT-B) with conventional interactive segmentation algorithms using the NoC metrics. Some methods employ large transformers including ViT-B, ViT-L and ViT-H. The best performances by the recent methods are obtained using the ViT-H backbone. However, ViT-H suffers from high computational costs due to its large number of parameters. For this reason, we compare the proposed algorithm with the state-of-the-art methods by employing ViT-B. Early algorithms were trained on the SBD dataset only, while recent methods are trained on SBD and COCO + LVIS, respectively. We report the results of MFP under both training settings following the recent methods.

Table~\ref{table:comparative_assessment_sbd} compares the results using the SBD training data. For the methods \cite{sofiiuk2022reviving,liu2022pseudoclick,chen2022focalclick, du2023efficient,lin2022focuscut} that presented multiple versions of their algorithms, we report their best NoC scores. Looking at the table results, we see that MFP outperforms the existing algorithms significantly. Note that MFP is the only algorithm that achieves a mean IoU of 85\% with less than four clicks for all the datasets. Also, MFP yields the best results in 9 out of 12 tests, and the second-best results in the remaining three tests.

Figure~\ref{fig:qualitative_comparison} compares qualitative segmentation results of the proposed MFP with those of FocalClick \cite{chen2022focalclick}, EMC-Click \cite{du2023efficient}, and SimpleClick \cite{liu2023simpleclick}. We can see that at the same number of clicks, MFP provides more accurate segmentation results. We also visualize previous probability maps $\tilde{P}^{t-1}$ in rows (b) of Figure~\ref{fig:qualitative_comparison}. Unlike the existing algorithms, whose segmentation masks are inaccurate in regions far from the clicks (\eg the bicycle wheels), MFP manages to make correct predictions in those regions. Specifically, the previous probability maps of both SimpleClick and MFP contain information about the wheel shapes, but only MFP can segment out the wheels. From this observation, we believe that the proposed MFP algorithm makes better use of the information in previous masks.

Table~\ref{table:comparative_assessment_coco} compares the results using the COCO + LVIS training data. Out of 12 NoC scores compared, MFP outperforms the existing algorithms in seven tests and ranks second in the remaining five tests. This demonstrates that the proposed MFP algorithm exceeds or shows comparable results to the state-of-the-art methods.

\vspace*{0.15cm}
\noindent\textbf{Comparison of IoU \& AUC:}
Figure~\ref{fig:IoUs} compares the proposed MFP in terms of mean IoU ratios with four comparable algorithms trained on the SBD dataset: CDNet \cite{chen2021conditional}, RITM \cite{sofiiuk2022reviving}, GPCIS \cite{zhou2023interactive}, SimpleClick \cite{liu2023simpleclick}. We see that MFP generally achieves higher mean IoU ratios with the same number of clicks than the conventional algorithms. To numerically prove the superiority of MFP, we also report the AUC scores. MFP shows the highest AUC scores on all four datasets.

\vspace*{0.15cm}
\noindent\textbf{Comparison using identical backbones:}
Using stronger backbones greatly impacts the performance of interactive segmentation algorithms. Thus, to make fair comparisons with existing methods, we test the proposed MFP algorithm using three different backbones: ResNet-34 \cite{he2016deep}, HRNet-18 \cite{wang2020deep}, and ViT-B \cite{dosovitskiy2021an}. We choose ViT-B as the main backbone following the state-of-the-art methods in  \cite{liu2023simpleclick, li2023interactive}. As ViT-B is a heavy network, we also choose relatively lighter backbones ResNet-34 and HRNet-18, which are employed by many other interactive segmentation methods. In Table~\ref{table:backbone-comparisons}, we compare the three versions of MFP with conventional algorithms that use the same backbone networks. Out of total 36 settings compared, MFP shows superiour results in 28 settings. Even for the cases where MFP does not achieve the highest scores, MFP yields comparable performance to the best results with margins less than 0.1. This demonstrates the effectiveness and generality of the proposed MFP framework.

\subsection{Ablation Study}
We conduct ablation studies to verify the efficacy of each component of the proposed algorithm, using the SBD training data. For evaluation, we choose the DAVIS dataset \cite{perazzi2016benchmark}. DAVIS has high-quality annotations and covers many different scenarios, thus we found the DAVIS results more convincing as compared with other datasets. The results are presented in Table~\ref{table:ablation}. We first implement a baseline model (Method~\RomNum{1}) based on ViT-B. Method~\RomNum{1} does not take modulated probability maps as additional input and follows the same network architecture and training scheme as the previous algorithm \cite{liu2023simpleclick}. In method~\RomNum{2}, we explore the effect of the late fusion strategy, by excluding the fusion layer. For method~\RomNum{3}, we follow the training procedure in \cite{sofiiuk2022reviving}, instead of the recursive training. When random sampling simulates clicks, since there is no order in the clicks, we randomly take any click and consider it as the current click to perform the probability map modulation. Method~\RomNum{4} employs all components proposed in this work, so it is the same as the proposed algorithm.

\begin{table}[t]\centering
    \renewcommand{\arraystretch}{1.5}
    \caption
    {
        Ablation studies of the proposed MFP algorithm conducted on the DAVIS dataset \cite{perazzi2016benchmark}.
    }
    \vspace*{-0.15cm}
    \resizebox{1.0\linewidth}{!}{
    \begin{tabular}[t]{+L{0.5cm}^C{1.5cm}^C{1.5cm}^C{1.5cm}^C{1.2cm}^C{1.2cm}^C{1.2cm}}
    \toprule
    & $\tilde{P}^{t-1}$ & Late fusion & Recursive training & NoC@85 & NoC@90 & NoC@95 \\
    \midrule
         \RomNum{1}. & & &                                  & 4.10 & 5.60 & 11.58 \\
         \RomNum{2}. & \checkmark &            & \checkmark & 4.05 & 5.30 & 11.54 \\
         \RomNum{3}. & \checkmark & \checkmark &            & 3.98 & 5.35 & 11.79\\
         \RomNum{4}. & \checkmark & \checkmark & \checkmark & 3.92 & 5.32 & 11.27 \\

    \bottomrule
    \end{tabular}}
    \vspace*{-0.5cm}
    \label{table:ablation}
\end{table}

Looking into the results in Table~\ref{table:ablation}, we can see that method \RomNum{4}, which employs all components, yields the best performance. This indicates that all components contribute to performance improvements. Comparing the results between pairs of methods, we find that the performance gain from method \RomNum{2} to \RomNum{4} is conspicuous for NoC@85, while that from method \RomNum{3} to \RomNum{4} stands out for NoC@95. From this observation, we deduce that fusing probability information in a late layer has great impact for quickly obtaining coarse predictions, while the recursive training scheme aids in fine-level tuning for higher accuracy.

\section{Conclusions}
\label{sec:conclusion}
A novel click-based interactive segmentation framework, called MFP, which fully exploits previous probability maps was proposed in this paper. First, MFP modulates previous probability maps based on click information to obtain better representations of user-specified objects. Then, it propagates the additional information to the segmentation network, which was designed by extending the existing interactive segmentation framework. Experimental results demonstrated that MFP outperforms conventional algorithms when identical backbones are employed.

\vspace*{-0.1cm}
\section*{Acknowledgements}
This work was supported by the NRF grants funded by the Korea government (MSIT) (No.~NRF-2021R1A4A1031864 and No.~NRF-2022R1A2B5B03002310), and by the IITP grant funded by the Korea government (MSIT) (No.~2021-0-02068, Artificial Intelligence Innovation Hub).

{
    \small
    \bibliographystyle{ieeenat_fullname}
    \bibliography{main}
}


\end{document}